\pdfoutput=1

\documentclass[11pt]{article}

\usepackage[preprint]{acl}

\usepackage{times}
\usepackage{latexsym}

\usepackage[T1]{fontenc}

\usepackage[utf8]{inputenc}

\usepackage{microtype}

\usepackage{inconsolata}

\usepackage{graphicx}

\usepackage{natbib}
\usepackage{amsmath}
\usepackage{multirow}
\usepackage{array}
\usepackage{booktabs}
\usepackage{lipsum}
\usepackage[hypcap=true]{caption}
\usepackage{subcaption}
\usepackage{amssymb}
\DeclareMathOperator*{\argmax}{arg\,max}

\title{RankAdaptor: Hierarchical Rank Allocation for Efficient Fine-Tuning Pruned LLMs via Performance Model}

\author{
    \textbf{Changhai Zhou\footnotemark[1]\textsuperscript{1,2}},
    \textbf{Shijie Han\thanks{Equal contribution}\textsuperscript{1,3}},
    \textbf{Lining Yang\textsuperscript{4}},
    \\
    \textbf{Yuhua Zhou\textsuperscript{5}},
    \textbf{Xu Cheng\textsuperscript{6}},
    \textbf{Yibin Wang\textsuperscript{1,2}},
    \textbf{Hongguang Li\thanks{Corresponding author: \href{mailto:harvey2@mail.ustc.edu.cn}{harvey2@mail.ustc.edu.cn}}\textsuperscript{1}}
    \\
    \\
    \textsuperscript{1}JF SmartInvest Holdings Ltd,
    \textsuperscript{2}Fudan University,
    \textsuperscript{3}Columbia University,
    \\
    \textsuperscript{4}King's College London,
    \textsuperscript{5}Zhejiang University,
    \textsuperscript{6}Wuhan University
}

\begin{document}
\maketitle

\begin{abstract}
The efficient compression of large language models (LLMs) has become increasingly popular. However, recovering the performance of compressed LLMs remains a major challenge. The current practice in LLM compression entails the implementation of structural pruning, complemented by a recovery phase that leverages the Low-Rank Adaptation (LoRA) algorithm. Structural pruning's uneven modification of model architecture, coupled with standard LoRA's fixed configuration allocation across layers in an online pipeline, leads to suboptimal performance in various downstream tasks for pruned models. To address this challenge, we introduce RankAdaptor, a hierarchical rank allocation method that enables efficient fine-tuning of pruned LLMs according to layerwise specific recovery requirements. We employ a performance model that conducts offline meta-learning and online incremental learning to explore optimal rank values for each layer. Comprehensive experiments on popular benchmarks show that RankAdaptor consistently outperforms state-of-the-art methods across a variety of pruning settings and LLM architectures, with improvements ranging from 0.7\% to 5.5\%.
\end{abstract}
\section{Introduction}
In recent years, large language models (LLMs) have provided innovative solutions across various natural language processing (NLP) tasks, such as machine translation \cite{zhang2023prompting, sato2020vocabulary, aycock2024topic}, sentiment analysis \cite{zhang2023enhancing, deng2023llms}, and speech recognition \cite{min2023exploring, fathullah2024prompting}. 

However, the exceptional performance of LLMs comes at the cost of a massive number of parameters and high-end hardware resources. Current compression techniques like pruning \cite{ma2023llm, xia2023sheared, santacroce2023matters,frantar2023sparsegpt}, quantization \cite{shao2023omniquant, lee2023enhancing}, and distillation \cite{gu2023minillm, tan2023gkd} have been explored. Compressed LLMs typically require fine-tuning to recover their original performance. Therefore, designing an efficient algorithm for compressed LLMs to achieve optimal performance on downstream tasks has become a pioneering direction.

Among compression techniques, structural pruning is a popular one that removes redundant weight connections to reduce model size and computational requirements. It primarily involves two stages: (1) pruning based on architectural importance and (2) recovery using efficient fine-tuning. While research has primarily focused on the initial pruning stage, the equally crucial recovery stage has been understudied. Existing approaches often rely on standard LoRA \cite{hu2021lora} for recovering pruned models, applying a general rank configuration across all layers. However, this approach overlooks the inherent structural irregularities introduced by pruning. Therefore, a one-size-fits-all rank configuration may not optimally meet the unique needs of each layer, potentially affecting downstream performance.

\begin{figure*}[tb!]
\centering
\includegraphics[width=0.9\linewidth]{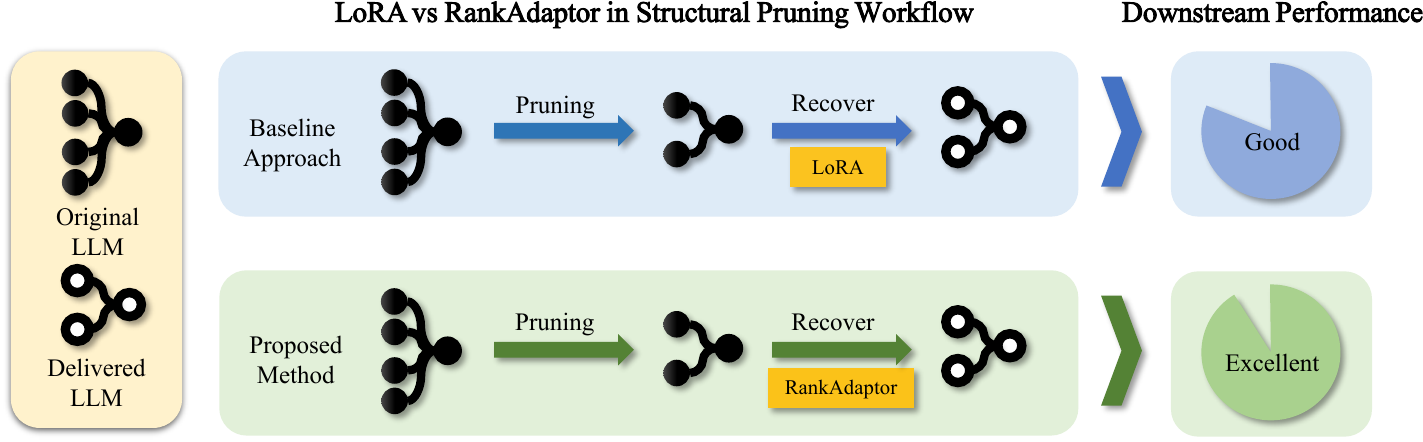}
\caption{Illustration of the process of pruning and recovery. The baseline approach is detailed in Section \ref{sec: bg}, and the proposed method is described in Section \ref{sec: method}.}
\label{fig: LLM-Pruner Workflow}
\end{figure*}

\begin{table}[tb!]
\centering
\caption{Zero-shot performance comparison between AdaLoRA \cite{zhang2023adaptive} and LoRA \cite{hu2021lora}. 'Bold' indicates better performance. 'Avg Diff' represents the average performance difference between AdaLoRA and LoRA across all benchmarks. The results are reported in percentage (\%).}
\label{tab: adalora_lora}
\resizebox{\linewidth}{!}{%
\begin{tabular}{@{}lccccccccc@{}}
\toprule
Method  & BoolQ          & PIQA           & HellaS         & WinoG          & ARC-e          & ARC-c          & OBQA           & Avg Diff \\
\midrule
AdaLoRA & 61.90          & 76.31          & 67.82          & 62.57          & \textbf{64.02} & 36.70          & 40.40          & \multirow{2}{*}{\textbf{0.56}} \\
LoRA    & \textbf{63.30} & \textbf{76.82} & \textbf{68.68} & \textbf{63.38} & 63.76          & \textbf{37.11} & \textbf{40.60} & \\
\bottomrule
\end{tabular}%
}
\end{table}
Among the various LoRA variants, AdaLoRA \cite{zhang2023adaptive} proposes an importance-based adaptive rank allocation method. It dynamically adjusts the rank configurations for each layer by continuously estimating the model structure's importance through SVD parameterization during fine-tuning. Despite AdaLoRA's proven advantages in fine-tuning original LLMs, its effectiveness in recovering accuracy for irregularly pruned models falls short of standard LoRA. We conducted experiments using a 20\% pruned LLaMA-7B model, comparing LoRA with rank=8 and AdaLoRA for performance recovery. As shown in Table \ref{tab: adalora_lora}, AdaLoRA's performance across seven tasks averages 0.56\% lower than LoRA. The suboptimal performance of AdaLoRA may be attributed to its difficulty in identifying the highly complex structures of pruned models during its online rank adjustment process. Given this observation, we propose adopting a static rank allocation strategy for each layer during the recovery stage of pruned models.

To achieve this goal, we propose RankAdaptor, an algorithm that leverages a performance model to statically determine the optimal rank configuration for each layer in the recovery stage for the pruned models. Our contributions are as follows:
\begin{enumerate}
    \item We point out a critical bottleneck in pruned LLM recovery: existing fine-tuning approaches fail to address the unique requirements of pruned models' complex structures.
    \item We introduce RankAdaptor, a tailored fine-tuning strategy specifically designed for recovering pruned models. Our approach employs a performance model that combines offline meta-learning with online incremental learning to efficiently explore optimal hierarchical rank configuration.
    \item Extensive experimentation has demonstrated that RankAdaptor consistently outperforms the state-of-the-art method across a range of pruning configurations and LLM architectures, with improvements ranging from 0.7\% to 5.5\%.
\end{enumerate} 
\section{Background and Motivation}\label{sec: bg}
There are numerous compression methods for LLMs and our work focuses on pruning. The entire pruning process for LLMs primarily consists of two main stages: (1) Pruning based on structural importance, and (2) Recovery using efficient fine-tuning, typically with LoRA.

\paragraph{Pruning Stage.}
This stage involves identifying and removing less important structures within the LLM. The process begins by establishing structural dependencies among neurons. A neuron $N_j$ is considered dependent on neuron $N_i$ if:
\begin{equation}
\begin{aligned}
     N_j \in \operatorname{Out}(N_i) \wedge \operatorname{Deg}^-(N_j) = 1 \\
     \Rightarrow N_j \text{ is dependent on } N_i
\end{aligned}
\end{equation}
where $\operatorname{Deg}^-(N_j)$ represents the in-degree of $N_j$. This dependency means that if $N_i$ is pruned, $N_j$ must also be pruned. The process identifies and groups dependent neurons, forming clusters of interconnected structures.

The importance of each group is then assessed using a Taylor expansion-based formula:
\begin{equation}
\begin{split}
    I_{\mathbf{W_i^k}} \approx \bigg| \frac{\partial \mathcal{L}}{\partial \mathbf{W_i^k}} \mathbf{W_i^k}
    &- \frac{1}{2} \sum_{j=1}^N \left(\frac{\partial \mathcal{L}_j}{\partial \mathbf{W_i^k}} \mathbf{W_i^k}\right)^2 \\
    &+ \mathcal{O}(\| \mathbf{W_i^k} \|^3) \bigg|
\end{split}
\label{eq: element_final_taylor}
\end{equation}
where $\mathbf{W_i^k}$ is the k-th parameter in structure $\mathbf{W_i}$, and $\mathcal{L}$ is the loss for next-token predictions. The groups are then ranked by importance, and those with lower significance are pruned based on a predefined ratio.

\paragraph{Recovery Stage.}
After pruning, the model's performance is recovered using efficient fine-tuning, typically through LoRA. In LoRA, for each layer of the pruned LLM, the weight update matrix $\Delta \mathbf{W}$ is decomposed into the product of two low-rank matrices $\mathbf{A}$ and $\mathbf{B}$:
\begin{equation}
\begin{split}
f(x) &= (\mathbf{W} + \Delta \mathbf{W})\mathbf{X} + \mathbf{b} \\
     &= (\mathbf{WX} + \mathbf{b}) + \mathbf{(AB)X}
\end{split}
\end{equation}
where $\mathbf{A} \in \mathbb{R}^{d \times r}$ and $\mathbf{B} \in \mathbb{R}^{r \times d}$, with $r$ being the rank, typically fixed across all layers. Only $\Delta \mathbf{W}$ (i.e., $\mathbf{AB}$) is updated during fine-tuning, while the original weight matrix $\mathbf{W}$ remains frozen. This approach significantly reduces the number of trainable parameters and computational cost from $d^2$ to $2dr$, as $r$ is usually much smaller than $d$.

\paragraph{A Motivating Example.}\label{2.3}
The uneven distribution of importance within LLMs' internal architecture \cite{zhang2023adaptive}, coupled with importance-based pruning criteria, leads to non-uniform pruning across layers. This results in a highly complex structure for the pruned LLM. While standard LoRA with fixed rank configurations offers some recovery, it falls short in addressing the specific recovery needs of differently pruned layers, leading to suboptimal performance.
\begin{figure}
    \centering
    \includegraphics[width=\linewidth]{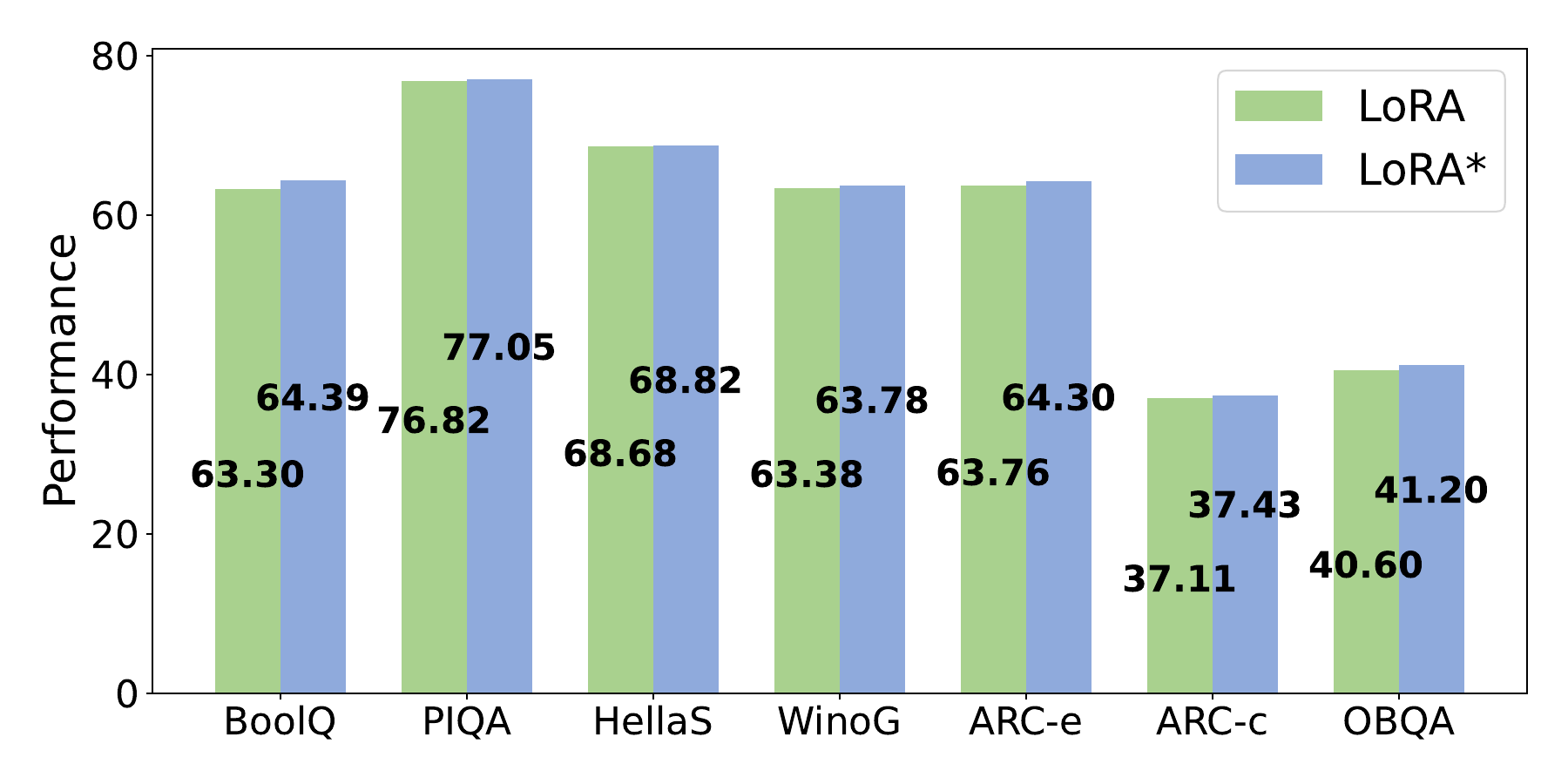}
    \caption{ 
    Performance of benchmarks for the different fine-tuning configurations. LoRA denotes using fixed ranks for different layers, whereas LoRA$^*$ indicates using different rank configurations. The results are reported in percentage (\%).}
    \label{fig: motivation}
\end{figure}

Studies~\cite{vaswani2017attention, zhao2024explainability} indicate that bottom layers in LLMs capture more semantic information, making them more powerful. Based on this insight, we explore two approaches for an LLaMA-7B model with a 20\% pruning ratio. Standard LoRA applies a fixed rank configuration of 8 across all layers, and LoRA$^{*}$ assigns increasing rank configuration from the bottom to the top layers. Specifically, in LoRA$^{*}$, layers 1--8 use rank 4, layers 9--16 use rank 6, layers 17--24 use rank 10, and layers 25--32 use rank 12.

The exploration in Figure \ref{fig: motivation} demonstrates the efficacy of different rank allocations for recovering pruned LLMs. LoRA with fixed rank configurations demonstrates general accuracy since its inability to meet the specific recovery requirements of different layers results in performance inferior to LoRA$^*$ across most tasks. LoRA$^*$ achieves superior recovery performance by gradually adjusting rank configurations for each layer during fine-tuning. 

Through this motivating example, we have demonstrated the effectiveness of using different rank configurations across layers for fine-tuning pruned models. However, determining the optimal rank configuration allocation for each layer remains a challenge. In the next section, we propose a method that uses a performance model to find the best combination of rank configuration for each layer.
\section{Methodology}\label{sec: method}
\begin{figure}
    \centering
    \includegraphics[width=\linewidth]{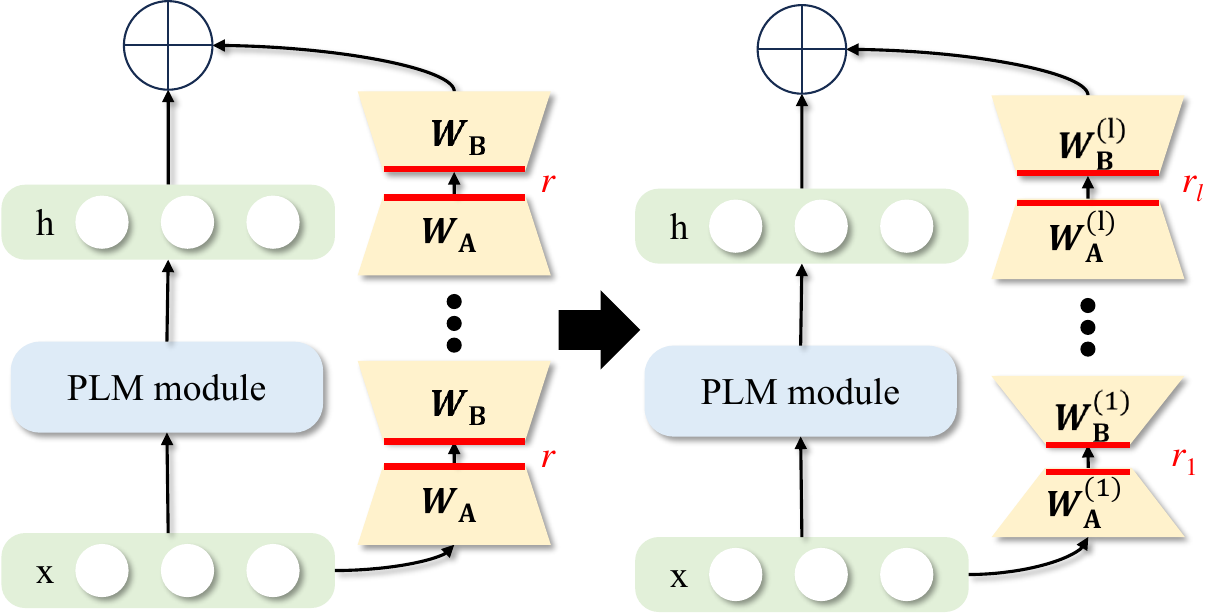}
    \caption{Hierarchical weight matrices decomposition: same rank in LoRA (left) versus hierarchical different ranks in RankAdaptor (right).}
\label{fig: diff_lora}
\end{figure}

In this section, we propose RankAdaptor, a hierarchical rank allocation tailored for fine-tuning pruned LLMs. The visible comparison from LoRA can be found in Figure \ref{fig: diff_lora}.

\subsection{Problem Definition}\label{3.1}
As mentioned in Section~\ref{2.3}, using a globally fixed rank $r$ in LoRA during the recovery stage can lead to suboptimal performance. While AdaLoRA attempts to dynamically adjust rank configurations for different layers during recovery, it is designed for unpruned models and proves unsuitable for pruned models. In addition, testing all combinations in the solution space $S$ is impractical for LLMs due to the vast number of layers and potential rank configurations. With $n$ rank candidates and $l$ layers, the number of combinations $n^l$ becomes astronomically large, rendering exhaustive evaluation unfeasible.

\paragraph{Problem Formulation.}
Given a pruned model $PL$ and a collection of rank configurations for all layers $R_{\text{H}}$, our overall objective is to identify the optimal rank set $R_{\text{H}}^*$ that maximizes the recovery performance. This can be expressed as:
\begin{equation}
    R_{\text{H}}^* = \argmax_{R_{\text{H}} \in S} \mathcal{P}(recover(PL, R_{\text{H}})),
\end{equation}
where $\mathcal{P}(recover(PL, R_{\text{H}}))$ represents the actual performance of recovering $PL$ with $R_{\text{H}}$. However, finding $R_{\text{H}}^*$ directly poses significant challenges. To address these, we propose an efficient method utilizing a predictive performance model, detailed in Section~\ref{3.2}. 

\paragraph{Training Objective.}Let $\mathcal{Q}(R_{\text{H}})$ denote the recovered performance predicted by our performance model, approximating $\mathcal{P}(recover(PL, R_{\text{H}}))$. If we can develop a model that takes $R_{\text{H}}$ as input and directly produces a performance prediction closely approximating the actual performance, we can efficiently explore a wide range of $R_{\text{H}}$ configurations at minimal cost and select the optimal one. 

Consequently, we can formulate our training objective to focus on training a reliable performance prediction model, which means minimizing the discrepancy between the actual performance and the predicted performance:
\begin{equation}
    \min \left[ \mathcal{P}(recover(PL, R_{\text{H}})) - \mathcal{Q}(R_{\text{H}}) \right]^2.
\end{equation}

\begin{figure*}[tb!]
    \centering
    \includegraphics[width=\linewidth]{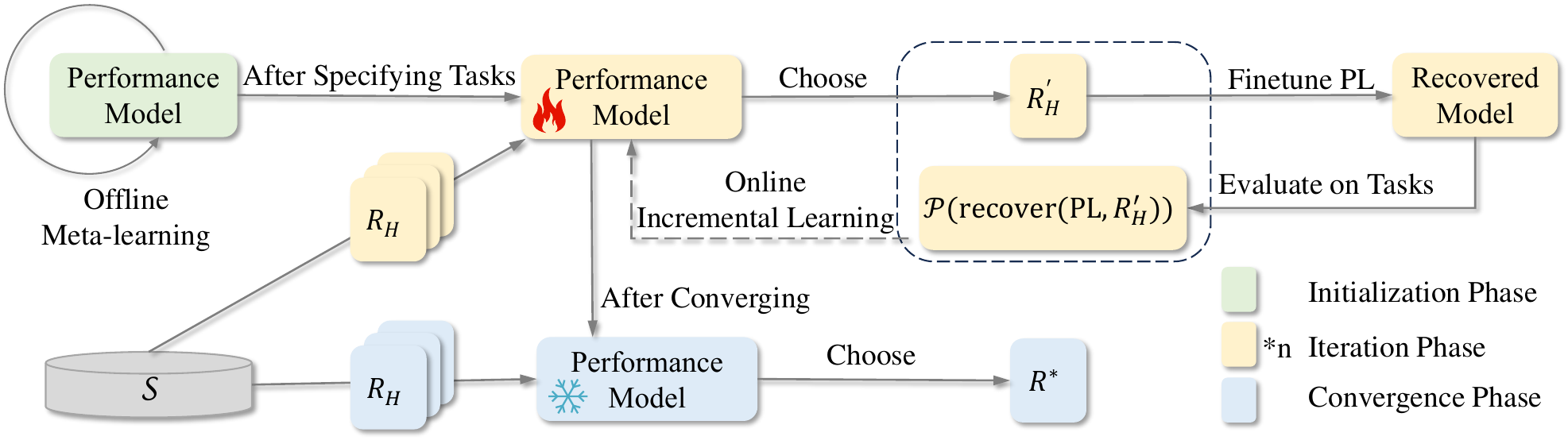}
    \caption{
        RankAdaptor Workflow: Through three phases (Initialization-Iteration-Convergence), find the optimal hierarchical rank configuration for recovering pruned LLM.
    }
    \label{fig: RankAdaptor}
\end{figure*}

\subsection{Performance Model.}\label{3.2}
A performance model is constructed to estimate the performance of the recovered model fine-tuned by $R_{\text{H}}$ on downstream tasks. Input $R_{\text{H}}$ to obtain $\mathcal{Q}(R_{\text{H}})$, which is a predicted configuration of $\mathcal{P}(recover(PL, R_{\text{H}}))$. 
\paragraph{Model Architecture}
Our performance model is inspired by the MLP architecture, featuring a simple and efficient structure that minimizes overhead during forward inference and backward propagation. We define the performance model as an MLP network comprising five fully connected layers. The input layer accepts $R_{\text{H}}$, representing the fine-tuning rank configurations of the pruned LLM as an $l$-dimensional vector, where $l$ is the number of layers in the LLM. Each hidden layer has a dimension $D_i$ where $i=1,2,3$, which can be adjusted as needed. The final output layer employs a linear activation function to generate a single scalar value representing the predicted performance score.

\paragraph{Model Integration.}
The performance model operates in two distinct phases. 
\begin{enumerate}
    \item \textbf{Offline meta-learning:} Before actual fine-tuning, we pre-train the performance model on multiple diverse datasets. This meta-learning approach endows the model with the ability to quickly adapt to new tasks and datasets, providing a foundation of generalized knowledge about rank configuration performance across various scenarios. 
    \item \textbf{Online incremental learning:} Once a specific downstream task is identified, the performance model is integrated into the RankAdaptor workflow. This phase enables rapid and accurate performance estimation for a large set of candidate rank configurations on the specific downstream task. By incrementally updating its knowledge based on task-specific data, the performance model refines its predictions to better align with the unique characteristics of the target task.
\end{enumerate}

\subsection{RankAdaptor}
\paragraph{Overview.}
Combined performance model, we propose RankAdaptor, a learning-based algorithm, as shown in Figure \ref{fig: RankAdaptor}, to allocate rank configuration for each layer. There are three important phases in our design: Initialization, Iteration, and Convergence. The first phase involves meta-learning pretraining of the performance model using multiple datasets. The function of his phase is to equip the model with fundamental learning abilities and generalization capabilities. The subsequent iterative phase focuses on incrementally enhancing the model's predictive power through continuous learning on specific tasks. In the final convergence stage, the performance model is utilized to predict performance for a large number of candidate rank configurations from the $S$. This process enables the selection of the optimal configuration that demonstrates superior performance on downstream tasks.

\paragraph{Initialization Phase.}
At the beginning of RankAdaptor, a lightweight performance model offline using meta-learning techniques is introduced. This process involves randomly selecting multiple configurations of $R_{\text{H}}$ from the solution space $S$, with a different set of $R_{\text{H}}$ being chosen for each step. The actual performance $\mathcal{P}(\text{recover}(PL, R_{\text{H}}))$ of these configurations across various datasets is used as a training set to update the performance model, serving as a foundation for subsequent optimization steps.

\paragraph{Iteration Phase.}
After the initialization phase, we enter the iteration phase. Here, many $R_{\text{H}}$ samples are randomly drawn from $S$ at each iteration. The performance model with the learning ability gained from the last phase predicts their corresponding performance $\mathcal{Q}(R_{\text{H}})$. Based on these predictions, we identify the optimal $R_{\text{H}}'$ for the current iteration, which is expected to yield the best accuracy on the downstream task. We then real fine-tune the $PL$ using $R_{\text{H}}'$ and evaluate its actual performance $\mathcal{P}(\text{recover}(PL, R_{\text{H}}'))$ on tasks. This performance data $(R_{\text{H}}', \mathcal{P}(\text{recover}(PL, R_{\text{H}}'))$ is fed back into the performance model, enabling continuous improvement of its predictive accuracy through successive iterations.

\paragraph{Convergence Phase.}
The process continues until the discrepancy between predicted performance $\mathcal{Q}(R_{\text{H}})$ and actual performance $\mathcal{P}(recover(PL, R_{\text{H}}))$ falls within a predetermined threshold. At this point, the performance model is considered converged. RankAdaptor can then efficiently identify the optimal $R_{\text{H}}^*$ from many $R_{\text{H}}$ in $S$ that maximizes the actual performance metrics of the pruned model.
\section{Experiments}

\subsection{Experimental Setup}
\label{4.1}
\noindent{\textbf{LLMs and Benchmarks.}}
To demonstrate the effectiveness of RankAdaptor, we test it on three open-source LLMs: LLaMA-7B\footnote{\url{https://huggingface.co/baffo32/decapoda-research-llama-7B-hf}} \cite{touvron2023llama}, LLaMA-13B\footnote{\url{https://huggingface.co/yahma/llama-13b-hf}} \cite{touvron2023llama} and Vicuna-7B\footnote{\url{https://huggingface.co/lmsys/vicuna-7b-v1.5}} \cite{zheng2024judging}. We conduct these LLMs on zero-shot classification tests for commonsense reasoning datasets, including BoolQ \cite{clark2019boolq}, PIQA \cite{bisk2020piqa}, HellaSwag \cite{zellers2019hellaswag}, WinoGrande \cite{sakaguchi2021winogrande}, ARC-easy \cite{clark2018think}, ARC-challenge \cite{clark2018think}, and OpenbookQA \cite{mihaylov2018can}. 

\paragraph{Baseline and Configuration.}
We employ both LLM-Pruner \cite{ma2023llm} and Shortened LLaMA (Taylor+) \cite{kimshortened} as pruning stage operations, which allow us to validate our method's effectiveness across different pruning strategies. We apply AdaLoRA \cite{zhang2023adaptive} and LoRA \cite{hu2021lora} as recovery methods compared with our RankAdaptor. 

Previous research has identified that specific layers of LLaMA-7B, Vicuna-7B, and LLaMA-13B are crucial to the models' architecture and should remain unpruned \cite{ma2023llm}. Thus, we prune only layers 5-30 of LLaMA-7B and Vicuna-7B, and layers 5-36 of LLaMA-13B to achieve the predefined global pruning rate. Specifically, we prune 25\%, 32\%, 38\%, and 63.5\% of the middle layers to attain global pruning rates of 20\%, 25\%, 30\%, and 50\%, respectively. For the unpruned layers, we maintain their rank configurations consistent with those of standard LoRA.

\begin{table*}[t]
    \centering
    \caption{Zero-shot performance of pruned LLaMA-7B with AdaLoRA, LoRA, and RankAdaptor recovery. 'Bold' indicates the best performance at each pruning rate. 'Avg' represents the average performance across all benchmarks. Specific rank configurations explored by RankAdaptor are listed in Appendix \ref{app:r}. Reported in percentage (\%).}
    \resizebox{0.95\linewidth}{!}{
        \begin{tabular}{cc|l|ccccccc|c}
            \toprule
            \multicolumn{2}{c|}{Pruning Stage} & Recover & BoolQ & PIQA & HellaS & WinoG & ARC-e & ARC-c & OBQA & Avg \\
            \midrule
            \multirow{12}{*}{\rotatebox[origin=c]{90}{LLM-Pruner}} 
            & \multirow{3}{*}{\parbox{1.8cm}{Rate = 20\%}} 
            & AdaLoRA & 61.90 & 76.31 & 67.82 & 62.57 & 64.02 & 36.70 & 40.40 & 58.53 \\
            & & LoRA  & 63.30 & 76.82 & 68.68 & 63.38 & 63.76 & 37.11 & 40.60 & 59.09 \\
            & & RankAdaptor & \textbf{67.34} & \textbf{77.31} & \textbf{69.07} & \textbf{64.17} & \textbf{65.36} & \textbf{37.80} & \textbf{41.60} & \textbf{60.38} \\
            \cmidrule{2-11}
            & \multirow{3}{*}{\parbox{1.8cm}{Rate = 25\%}} 
            & AdaLoRA & 60.31 & 75.82 & 64.57 & 61.30 & 61.88 & 35.10 & 39.00 & 56.85 \\
            & & LoRA  & 61.93 & 76.01 & \textbf{66.08} & 61.96 & 62.21 & 35.92 & 39.40 & 57.64 \\
            & & RankAdaptor & \textbf{67.43} & \textbf{76.06} & 65.97 & \textbf{64.40} & \textbf{62.63} & \textbf{36.77} & \textbf{40.40} & \textbf{59.09} \\
            \cmidrule{2-11}
            & \multirow{3}{*}{\parbox{1.8cm}{Rate = 30\%}} 
            & AdaLoRA & 59.85 & 73.30 & 61.53 & 60.12 & 59.67 & 33.21 & 38.80 & 55.21 \\
            & & LoRA  & 62.45 & 74.37 & 63.14 & 61.96 & 59.22 & 33.70 & 39.60 & 56.35 \\
            & & RankAdaptor & \textbf{66.21} & \textbf{75.19} & \textbf{63.61} & \textbf{63.14} & \textbf{60.10} & \textbf{34.64} & \textbf{40.20} & \textbf{57.58} \\
            \cmidrule{2-11}
            & \multirow{3}{*}{\parbox{1.8cm}{Rate = 50\%}} 
            & AdaLoRA & 38.25 & 67.86 & 42.80 & 48.32 & 42.20 & 26.31 & 32.80 & 42.65 \\
            & & LoRA  & 43.76 & 69.04 & 45.01 & 50.99 & \textbf{45.66} & \textbf{28.75} & 34.60 & 45.40 \\
            & & RankAdaptor & \textbf{51.65} & \textbf{69.48} & \textbf{45.03} & \textbf{51.93} & 45.20 & 28.41 & \textbf{35.00} & \textbf{46.67} \\
            \midrule
            \multirow{12}{*}{\rotatebox[origin=c]{90}{Shortened}} 
            & \multirow{3}{*}{\parbox{1.8cm}{Rate = 20\%}} 
            & AdaLoRA & 70.14 & 73.85 & 68.92 & 65.73 & 64.93 & 37.12 & 39.34 & 60.00 \\
            & & LoRA  & 71.82 & \textbf{75.31} & 70.50 & 67.36 & 64.40 & 38.60 & 40.80 & 61.26 \\
            & & RankAdaptor & \textbf{74.53} & 75.22 & \textbf{72.81} & \textbf{69.92} & \textbf{66.72} & \textbf{40.23} & \textbf{42.31} & \textbf{63.11} \\
            \cmidrule{2-11}
            & \multirow{3}{*}{\parbox{1.8cm}{Rate = 25\%}} 
            & AdaLoRA & 67.93 & 71.54 & 65.23 & 63.14 & 61.83 & 35.53 & 36.13 & 57.33 \\
            & & LoRA  & 69.67 & 73.20 & 66.85 & 64.50 & 63.32 & 37.02 & 37.40 & 58.85 \\
            & & RankAdaptor & \textbf{72.32} & \textbf{75.13} & \textbf{68.91} & \textbf{66.82} & \textbf{65.24} & \textbf{38.54} & \textbf{38.92} & \textbf{60.84} \\
            \cmidrule{2-11}
            & \multirow{3}{*}{\parbox{1.8cm}{Rate = 30\%}} 
            & AdaLoRA & 61.83 & 69.72 & 61.24 & 62.34 & 58.13 & 33.23 & 35.54 & 54.58 \\
            & & LoRA  & 63.58 & 71.21 & 62.75 & 63.80 & \textbf{59.42} & 34.50 & 36.80 & 56.01 \\
            & & RankAdaptor & \textbf{66.23} & \textbf{73.42} & \textbf{64.82} & \textbf{65.93} & 59.31 & \textbf{35.84} & \textbf{38.13} & \textbf{57.67} \\
            \cmidrule{2-11}
            & \multirow{3}{*}{\parbox{1.8cm}{Rate = 50\%}} 
            & AdaLoRA & 44.93 & 61.24 & 41.83 & 52.84 & 42.93 & 28.73 & 32.24 & 43.53 \\
            & & LoRA  & 46.52 & \textbf{62.76} & 43.12 & 54.37 & 44.10 & \textbf{30.07} & 33.50 & 44.92 \\
            & & RankAdaptor & \textbf{48.94} & 62.43 & \textbf{45.24} & \textbf{56.53} & \textbf{46.14} & 29.82 & \textbf{34.84} & \textbf{46.28} \\
            \bottomrule
        \end{tabular}
    }
    \label{tab: res-llama7b}
\end{table*}

\paragraph{Implementation Details.}
Our implementation utilizes PyTorch 2.1.2, Transformers 4.41.0, and PEFT 0.6.0 libraries, running on CUDA 12.4. The hardware setup consists of an NVIDIA A800 GPU with 80GB memory, operating on Ubuntu. The MLP dimensions for the inner layers of the performance model are set to (32-32-32-1), meaning each inner MLP consists of three hidden layers with 32 neurons and an output layer with a single neuron. Micro-batch size is configured to 16, which specifies the number of examples processed in each step of model training.

\paragraph{Rank Configuration Candidates and Solution Space.}
In standard LoRA, setting fixed rank configurations within the range of 2 to 16 achieves favorable model recovery. To ensure that the trainable parameter count of RankAdaptor remains at the same level as standard LoRA, the range of rank configurations in this experiment is set to $\left\{2, 4, 6, 8, 10, 12\right\}$. For LLaMA-7B and Vicuna-7B, which have 26 pruned layers, the size of the solution space is $6^{26}$. For LLaMA-13B, with 32 pruned layers, the size of the solution space is $6^{32}$. Different models follow the same calculation pattern.

\subsection{Main Results}
\paragraph{Analysis.}
We present the performance of the recovered LLM finetuned by AdaLoRA, LoRA, and RankAdaptor on each benchmark in Table \ref{tab: res-llama7b} below, and Tables \ref{tab: res-vicuna7b} and \ref{tab: res-llama13b} in the appendix. The performances of pruned LLM without recovery are listed in Appendix \ref{sec: wo_ft}. We have illustrated the specific configuration of the rank configuration explored by RankAdaptor in LLaMA-7B in Appendix \ref{app:r}.

RankAdaptor shows strong performance under different pruning strategies. Whether applied with the LLM-Pruner or the Shortened, RankAdaptor generally achieves the highest average scores across all benchmarks. This adaptability to different pruning approaches further highlights its robustness as a pruning recovery method.

At lower pruning rates (20-25\%), RankAdaptor shows remarkable effectiveness. For instance, in the LLaMA-13B model with a 20\% pruning rate, RankAdaptor achieves the highest scores in 6 out of 7 tasks. This trend continues with Vicuna-7B, where RankAdaptor leads in most tasks at 20\% and 25\% pruning rates. Even at higher pruning rates (30-50\%), where performance typically degrades more significantly, RankAdaptor maintains its edge. In the challenging scenario of 50\% pruning, RankAdaptor still manages to outperform other methods in most tasks for both LLaMA-13B and Vicuna-7B. 

Furthermore, RankAdaptor's effectiveness is also consistent across different types of tasks. In language-understanding tasks like BoolQ and HellaSwag, as well as in more reasoning-focused tasks like ARC and OBQA, RankAdaptor consistently achieves the best or near-best performance. This broad-spectrum effectiveness suggests that RankAdaptor is adept at preserving various aspects of language model capabilities, from basic comprehension to more complex reasoning.

In summary, the results offer substantial evidence in support of RankAdaptor's efficacy as a pruning recovery approach. Its consistent superiority across diverse scenarios illustrates that RankAdaptor is a highly effective technique, irrespective of model size, architecture, or pruning rate.

\paragraph{Generation Performance.}\label{sec: gen}
Complementing the evaluation of model performance on classification tasks in the experiments, we further investigate the generative capabilities of the recovered models. Notably, we conduct text generation tasks using LLaMA-7B and Vicuna-7B models recovered by LoRA and RankAdaptor at a 20\% pruning rate, as detailed in Appendix \ref{Generative}. The results are remarkably promising. For article continuation, the models recovered by RankAdaptor demonstrate superior coherence in their generated sentences. Similarly, when tasked with step listing, RankAdaptor-recovered LLMs produce clearer and more logical step sequences. These compelling comparative results are illustrated in Figure \ref{fig: gen1} and \ref{fig: gen2}, showcasing the potential of RankAdaptor in preserving and enhancing generative abilities during model compression and recovery.

\subsection{Ablation Study}
We prune LLaMA-7B with a 20\% global pruning rate and the RankAdaptor to recover. More details can be found in Table \ref{tab: Ablation}.

\begin{table*}[ht]
\centering
\caption{Ablation study results comparing performance across seven tasks using LLM-Pruner on LLaMA-7B. 'Bold' indicates better performance. The results are reported in percentage (\%).}
\label{tab: Ablation}
\resizebox{\linewidth}{!}
{
\begin{tabular}{c|cc|ccc|cc|ccc}
\toprule
\multirow{2}{*}{Benchmark} & \multicolumn{2}{c}{Sample Size} & \multicolumn{3}{c}{Micro-batch Size} & \multicolumn{2}{c}{Element-wise Importance} & \multicolumn{3}{c}{Setting of Performance Model} \\
\cmidrule(lr){2-3} \cmidrule(lr){4-6} \cmidrule(lr){7-8} \cmidrule(lr){9-11}
 & $N$=10 & $N$=50 & Micro-4 & Micro-8 & Micro-16 & Element\textsuperscript{1} & Element\textsuperscript{2} & Setting1 & Setting2 & Setting3 \\
\midrule
ARC-e       & 63.97 & \textbf{65.32} & 64.52 & 63.97 & \textbf{65.24} & \textbf{63.97} & 62.84 & 63.97 & 64.73 & 64.65 \\
ARC-c       & 37.29 & \textbf{37.71} & \textbf{38.65} & 37.29 & 37.54 & \textbf{37.29} & 36.77 & 37.29 & 36.60 & \textbf{37.54} \\
WinoG  & \textbf{63.61} & 63.14 & 62.04 & \textbf{63.61} & 63.14 & \textbf{63.61} & 63.22 & 63.61 & \textbf{63.46} & 63.06 \\
OBQA        & \textbf{39.80} & \textbf{41.00} & 40.00 & 39.80 & \textbf{40.80} & \textbf{39.80} & 39.80 & 39.80 & \textbf{40.80} & \textbf{40.80} \\
BoolQ       & \textbf{65.81} & 64.43 & \textbf{67.28} & 65.81 & 66.91 & 65.81 & \textbf{66.48} & \textbf{66.91} & 64.43 & 64.86 \\
PIQA        & 76.99 & \textbf{77.15} & 76.50 & \textbf{76.99} & 76.93 & \textbf{76.99} & 76.82 & 76.99 & 76.99 & \textbf{77.04} \\
HellaS   & \textbf{68.56} & 68.52 & 68.08 & 68.56 & \textbf{68.78} & \textbf{68.56} & 67.88 & 68.56 & 68.75 & \textbf{69.00} \\
\bottomrule
\end{tabular}
}
\end{table*}

\paragraph{Sample Size.}
We conduct ablation experiments to assess the impact of the estimated sample size during the pruning phase in LLM-Pruner. The larger the estimated sample size in the pruning operation, the better it can evaluate the importance of the model architecture and perform better pruning effects. So we compare performance with $N=10$ and $50$, and results demonstrate that increasing the sample size to $N=50$ leads to better outcomes. However, while a larger sample size ($N=50$) tends to improve performance for most tasks, there are instances where the smaller sample size ($N=10$) yields competitive results, such as in WinoGrande. This underscores the need for careful selection of sample size based on the specific requirements of the task.

\paragraph{Micro-batch Sizes.}
We finally assess the impact of different micro-batch sizes (4, 8, and 16) in fine-tuning process. The results indicate that larger micro-batch sizes can lead to better performance on certain tasks, though not universally across all benchmarks. 

\paragraph{Element-wise Importance.}
We further conduct tests on the LLM-Pruner's importance estimation techniques. The results compare the first-order (Element\textsuperscript{1}) and second-order (Element\textsuperscript{2}) Taylor approximations for evaluating the importance of each parameter, as described in Equation \ref{eq: element_final_taylor}. Our findings indicate that Element\textsuperscript{1} provides better performance than Element\textsuperscript{2} across most benchmarks. While higher-order derivatives may theoretically offer more precise adjustments, their complexity may outweigh the marginal performance gains observed in practice.

\paragraph{Setting of Performance Model.}
To investigate the impact of different inner MLP dimensions in the performance model, we test three configurations. The first setting consists of three hidden layers with 32 neurons each, followed by an output layer with a single neuron, abbreviated as 32-32-32-1. The other two configurations are 32-64-32-1 and 32-16-32-1, following the same notation. The results illustrate that varying dimensions of inner MLP layers have nuanced impacts on performance across different benchmarks. For inner MLP dimensions, Setting1 provides the highest performance on tasks such as ARC-e and BoolQ, while Setting3 shows competitive performance on PIQA and HellaSwag. 

\section{Related Work}
\subsection{Efficient Pruning of LLMs}
LLM-Pruner \cite{ma2023llm} employs structured pruning to remove non-essential interconnected structures by utilizing gradient information. This approach allows compressed models to restore good performance in multitasks with basic fine-tuning. \citet{xia2023sheared} introduces "Sheared LLaMA" to compress pre-trained LLMs. It employs dynamic batch loading to improve data efficiency during pruning and retraining. \citet{santacroce2023matters} presents Globally Unique Movement (GUM), a novel pruning technique that focuses on the sensitivity and uniqueness of LLMs' network components. GUM selects models' neurons that uniquely contribute to model output and are sensitive to loss changes to prune, thus maintaining high accuracy. SparseGPT \cite{frantar2023sparsegpt} transforms the pruning process into a series of large-scale sparse regression problems, which can be quickly solved through Hessian matrix inversion. It efficiently prunes large models to high sparsity in a single step while maintaining high accuracy. Wanda \cite{sun2023simple} prunes LLMs by selectively removing weights based on their sizes and input activations. It adaptively adjusts sparsity levels to achieve a reduction of more than half without sacrificing accuracy.

\subsection{Parameter Efficient Fine-Tuning}
\citet{houlsby2019parameter} introduce a transfer learning method that integrates adapter modules into pre-trained Transformer models. It can efficiently tackle various NLP tasks with few additional parameters and achieve performance similar to full fine-tuning. While the adapter takes a serial approach to integrating trainable components into pre-trained Transformer models, low-rank adaptation (LoRA) \cite{hu2021lora} presents a parallel method of infusing rank decomposition matrices into each layer of the model's architecture. Specifically, LoRA adds trainable matrices to each layer of the model and the pre-trained weights are kept the same. LoRA reduces the number of trainable parameters compared to fine-tuning the entire model, which makes model adaptation faster and less resource-intensive. LoRA-FA \cite{zhang2023lora} freezes the projection-down weight of the low-rank adaptation (LoRA) layers and only updates the projection-up weight to reduce the memory requirements for fine-tuning. \citet{zhang2023adaptive} have introduced AdaLoRA, which achieves excellent performance by parameterizing updates in SVD form and employing a novel importance metric to dynamically adjust hierarchical rank configurations during the fine-tuning process.
\section{Conclusion}

In this work, we present RankAdaptor, an innovative fine-tuning algorithm specifically designed to recover the performance of pruned LLMs. RankAdaptor employs a hierarchical fine-tuning strategy, incorporating a lightweight performance model to optimize rank configuration across different layers. This methodology effectively mitigates the drawbacks of the standard fixed-rank LoRA, which often results in suboptimal performance recovery due to the uneven architectural adjustments caused by structural pruning. Through extensive evaluations of multiple open-source LLMs and benchmark tasks, we demonstrate that RankAdaptor consistently outperforms the standard LoRA approach across various pruning scenarios. The introduction of RankAdaptor marks a significant progression in fine-tuning pruned LLMs. Its adaptive rank scheduling and end-to-end optimization lead to substantial enhancements over traditional techniques, positioning it as a promising tool for boosting the performance of pruned language models in diverse applications.

\section*{Limitations}
Our methodology delves into the optimization of fine-tuning procedures for pruned models, albeit the offline algorithm necessitates supplementary training data, thereby introducing some level of overhead. Subsequent endeavors will be directed towards refining this phase and integrating a broader range of quantization algorithms, with a focus on effectively fine-tuning the quantized model to achieve final performance and accuracy.

\bibliography{acl_latex}

\appendix
\clearpage
\section{RankAdaptor Appendix}
\begin{table*}[tb!]
    \centering
    \caption{Zero-shot performance of pruned Vicuna-7B with AdaLoRA, LoRA, and RankAdaptor recovery. 'Bold' indicates the best performance at each pruning rate. 'Avg' represents the average performance across all benchmarks. The results are reported in percentage (\%).}
    \resizebox{\linewidth}{!}{
        \begin{tabular}{cc|l|ccccccc|c}
            \toprule
            \multicolumn{2}{c|}{Pruning Stage} & Recover & BoolQ & PIQA & HellaS & WinoG & ARC-e & ARC-c & OBQA & Avg \\
            \midrule
            \multirow{12}{*}{\rotatebox[origin=c]{90}{LLM-Pruner}}  
            & \multirow{3}{*}{\parbox{1.8cm}{Rate = 20\%}} 
            & AdaLoRA & 55.43 & 75.22 & 65.31 & 60.89 & 64.96 & 35.61 & 38.10 & 56.50 \\
            & & LoRA  & 57.77 & \textbf{77.58} & 67.16 & 63.14 & 67.30 & 37.71 & 40.40 & 58.72 \\
            & & RankAdaptor & \textbf{61.19} & 77.15 & \textbf{67.32} & \textbf{63.85} & \textbf{67.68} & \textbf{38.05} & \textbf{41.20} & \textbf{59.49} \\
            \cmidrule{2-11}
            & \multirow{3}{*}{\parbox{1.8cm}{Rate = 25\%}} 
            & AdaLoRA & 48.24 & 72.89 & 62.25 & 59.18 & 61.58 & 33.57 & 38.30 & 53.72 \\
            & & LoRA  & 50.34 & 75.24 & 64.10 & 61.33 & \textbf{63.93} & 35.67 & 40.60 & 55.89 \\
            & & RankAdaptor & \textbf{58.50} & \textbf{76.17} & \textbf{64.23} & \textbf{61.96} & 63.30 & \textbf{36.01} & \textbf{42.00} & \textbf{57.45} \\
            \cmidrule{2-11}
            & \multirow{3}{*}{\parbox{1.8cm}{Rate = 30\%}} 
            & AdaLoRA & 56.46 & 71.87 & 58.85 & 58.27 & 56.66 & 31.69 & 36.50 & 52.90 \\
            & & LoRA  & \textbf{58.81} & 74.37 & 60.70 & \textbf{60.62} & 59.01 & 33.79 & 38.80 & 55.16 \\
            & & RankAdaptor & 57.58 & \textbf{75.57} & \textbf{61.63} & 60.22 & \textbf{60.94} & \textbf{34.81} & \textbf{39.00} & \textbf{55.68} \\
            \cmidrule{2-11}
            & \multirow{3}{*}{\parbox{1.8cm}{Rate = 50\%}} 
            & AdaLoRA & 57.21 & 64.52 & 41.83 & 49.66 & 46.05 & 24.35 & 31.70 & 45.05 \\
            & & LoRA  & \textbf{59.51} & 66.87 & 43.18 & \textbf{52.01} & 48.40 & 26.45 & 34.00 & 47.20 \\
            & & RankAdaptor & 59.91 & \textbf{67.46} & \textbf{43.50} & 52.41 & \textbf{48.70} & \textbf{27.65} & \textbf{35.80} & \textbf{47.92} \\
            \midrule
            \multirow{12}{*}{\rotatebox[origin=c]{90}{Shortened}} 
            & \multirow{3}{*}{\parbox{1.8cm}{Rate = 20\%}} 
            & AdaLoRA & 69.52 & 72.95 & 66.83 & 63.91 & 65.82 & 37.23 & 38.64 & 59.27 \\
            & & LoRA  & 71.82 & 74.32 & 68.45 & \textbf{67.62} & 67.35 & 38.50 & 39.80 & 61.12 \\
            & & RankAdaptor & \textbf{74.31} & \textbf{76.92} & \textbf{70.73} & 65.37 & \textbf{69.51} & \textbf{40.12} & \textbf{41.34} & \textbf{62.61} \\
            \cmidrule{2-11}
            & \multirow{3}{*}{\parbox{1.8cm}{Rate = 25\%}} 
            & AdaLoRA & 67.63 & 71.84 & 64.15 & 62.53 & 62.41 & 36.72 & 37.32 & 57.51 \\
            & & LoRA  & 69.85 & 73.21 & 65.72 & 64.08 & 63.90 & \textbf{39.32} & 38.50 & 59.23 \\
            & & RankAdaptor & \textbf{72.41} & \textbf{75.62} & \textbf{67.93} & \textbf{66.21} & \textbf{65.84} & 37.95 & \textbf{39.91} & \textbf{60.84} \\
            \cmidrule{2-11}
            & \multirow{3}{*}{\parbox{1.8cm}{Rate = 30\%}} 
            & AdaLoRA & 60.92 & 70.54 & 59.21 & 61.15 & 56.73 & 33.42 & 35.61 & 53.94 \\
            & & LoRA  & 62.75 & 71.93 & 60.68 & 62.70 & \textbf{60.32} & 34.50 & 36.80 & 55.67 \\
            & & RankAdaptor & \textbf{65.13} & \textbf{74.21} & \textbf{62.84} & \textbf{64.92} & 58.15 & \textbf{35.93} & \textbf{38.12} & \textbf{57.04} \\
            \cmidrule{2-11}
            & \multirow{3}{*}{\parbox{1.8cm}{Rate = 50\%}} 
            & AdaLoRA & 58.63 & 66.42 & 45.21 & 52.34 & 46.92 & 27.51 & 33.42 & 47.21 \\
            & & LoRA  & 60.37 & \textbf{69.73} & 46.50 & 53.76 & 48.25 & 28.72 & 34.60 & 48.85 \\
            & & RankAdaptor & \textbf{62.54} & 67.88 & \textbf{48.12} & \textbf{55.62} & \textbf{49.84} & \textbf{29.91} & \textbf{35.83} & \textbf{49.96} \\
            \bottomrule
        \end{tabular}
    }
    \label{tab: res-vicuna7b}
\end{table*}
\begin{table*}[tb!]
    \centering
    \caption{Zero-shot performance of pruned LLaMA-13B with AdaLoRA, LoRA, and RankAdaptor recovery. 'Bold' indicates the best performance at each pruning rate. 'Avg' represents the average performance across all benchmarks. The results are reported in percentage (\%).}
    \resizebox{0.85\linewidth}{!}{
        \begin{tabular}{cc|l|ccccccc|c}
            \toprule
            \multicolumn{2}{c|}{Pruning Stage} & Recover & BoolQ & PIQA & HellaS & WinoG & ARC-e & ARC-c & OBQA & Avg \\
            \midrule
            \multirow{3}{*}{\rotatebox[origin=c]{90}{LLM-}\rotatebox[origin=c]{90}{Pruner}}  
            & \multirow{3}{*}{\parbox{1.8cm}{Rate = 50\%}} 
            & AdaLoRA & 59.63 & 69.03 & 51.61 & 51.24 & 50.76 & 28.25 & 35.70 & 49.46 \\
            & & LoRA  & 61.93 & 71.38 & \textbf{53.36} & 53.59 & 53.11 & 29.95 & 38.00 & 51.62 \\
            & & RankAdaptor & \textbf{62.05} & \textbf{71.71} & 53.33 & \textbf{54.22} & \textbf{53.20} & \textbf{30.89} & \textbf{39.40} & \textbf{52.11} \\
            \midrule
            \multirow{3}{*}{\rotatebox[origin=c]{90}{Shortened}} 
            & \multirow{3}{*}{\parbox{1.8cm}{Rate = 50\%}} 
            & AdaLoRA & 71.53 & 70.92 & 54.32 & 52.83 & 50.94 & 31.62 & 39.21 & 53.05 \\
            & & LoRA  & 73.75 & \textbf{73.81} & 55.96 & 54.41 & 52.35 & 32.81 & 40.60 & 54.81 \\
            & & RankAdaptor & \textbf{75.92} & 72.64 & \textbf{57.83} & \textbf{56.23} & \textbf{54.12} & \textbf{33.96} & \textbf{41.92} & \textbf{56.09} \\
            \bottomrule
        \end{tabular}
    }
    \label{tab: res-llama13b}
\end{table*}
\subsection{Performance of Different Model Varients.}
We list the performance of the Vicuna-7B in Table \ref{tab: res-vicuna7b} and LLaMA-13B in Table \ref{tab: res-llama13b}. These results demonstrate the remarkable versatility and effectiveness of our RankAdaptor method across various dimensions. Firstly, the method exhibits consistent superior performance across different model architectures, as evidenced by its effectiveness on both Vicuna-7B and LLaMA-13B models. This cross-model applicability underscores the scalability and adaptability of our approach. Secondly, RankAdaptor demonstrates robustness to varying pruning rates, consistently outperforming AdaLoRA and LoRA across a wide range of pruning rates from 20\% to 50\%. This resilience indicates that our method maintains its effectiveness even under aggressive pruning scenarios. Thirdly, the superior performance of RankAdaptor is observed across diverse downstream tasks, including BoolQ, PIQA, HellaSwag, WinoGrande, ARC-easy, ARC-challenge, and OBQA, highlighting its task-agnostic nature. Finally, the method's efficacy is proven under different pruning strategies, namely LLM-Pruner and Shortened, further emphasizing its flexibility. In almost all scenarios, RankAdaptor achieves the highest average performance, often with significant margins, demonstrating its potential as a universal solution for recovering pruned language models across various configurations and applications.

\subsection{Performance of Pruned LLM without Recovery}\label{sec: wo_ft}
To demonstrate the critical importance of a recovery phase in achieving optimal performance for pruned models, we conducted extensive benchmark tests on LLaMA-7B models that had undergone only the pruning stage. The results, presented in Table \ref{tab: res-llama7b-wo}, offer compelling evidence when compared with those in Table \ref{tab: res-llama7b}. This comparison reveals a stark contrast in performance between models with and without recovery. Models that have not undergone recovery exhibit significantly diminished performance across all evaluated tasks, highlighting the potential loss of crucial learned representations during the pruning process. In contrast, models that have been subjected to a recovery process, regardless of the specific fine-tuning method employed (be it AdaLoRA, LoRA, or our proposed RankAdaptor), demonstrate substantial performance improvements across nearly all tasks. Furthermore, while all recovery methods show improvements, the degree of enhancement varies, with our proposed RankAdaptor method consistently achieving superior results. These findings emphasize that the recovery phase should not be considered an optional step but rather an integral component of the model compression process.

\begin{table*}[t]
    \centering
    \caption{Zero-shot performance of pruned LLaMA-7B without recovery. 'Avg' represents the average performance across all benchmarks. The results are reported in percentage (\%).}
    \resizebox{0.85\linewidth}{!}{
        \begin{tabular}{cc|ccccccc|c}
            \toprule
            \multicolumn{2}{c|}{Pruning Stage} & BoolQ & PIQA & HellaS & WinoG & ARC-e & ARC-c & OBQA & Avg \\
            \midrule
            \multirow{4}{*}{\rotatebox[origin=c]{90}{LLM-Pruner}}
            & \multirow{1}{*}{\parbox{1.8cm}{Rate = 20\%}} & 56.94 & 75.73 & 66.83 & 60.06 & 60.94 & 36.43 & 39.80 & 56.68 \\
            \cmidrule{2-10}
            & \multirow{1}{*}{\parbox{1.8cm}{Rate = 25\%}} & 59.94 & 73.23 & 62.35 & 58.80 & 55.81 & 34.90 & 39.40 & 54.92 \\
            \cmidrule{2-10}
            & \multirow{1}{*}{\parbox{1.8cm}{Rate = 30\%}} & 58.96 & 71.22 & 58.10 & 58.88 & 52.19 & 32.34 & 38.40 & 52.87 \\
            \cmidrule{2-10}
            & \multirow{1}{*}{\parbox{1.8cm}{Rate = 50\%}} & 57.98 & 60.94 & 34.35 & 52.25 & 31.82 & 27.30 & 35.80 & 42.92 \\
            \midrule
            \multirow{4}{*}{\rotatebox[origin=c]{90}{Shortened}} 
            & \multirow{1}{*}{\parbox{1.8cm}{Rate = 20\%}} & 62.65 & 77.24 & 68.83 & 61.26 & 62.76 & 37.52 & 40.99 & 58.75 \\
            \cmidrule{2-10}
            & \multirow{1}{*}{\parbox{1.8cm}{Rate = 25\%}} & 61.74 & 74.69 & 64.22 & 60.56 & 57.48 & 35.94 & 40.58 & 56.46 \\
            \cmidrule{2-10}
            & \multirow{1}{*}{\parbox{1.8cm}{Rate = 30\%}} & 60.73 & 74.64 & 59.84 & 59.64 & 53.75 & 33.31 & 39.55 & 54.49 \\
            \cmidrule{2-10}
            & \multirow{1}{*}{\parbox{1.8cm}{Rate = 50\%}} & 59.72 & 62.16 & 35.38 & 53.81 & 32.77 & 28.12 & 36.87 & 44.12 \\
            \bottomrule
        \end{tabular}
    }
    \label{tab: res-llama7b-wo}
\end{table*}

\subsection{Specific Rank Configuration Allocation} \label{app:r}
\begin{table*}[ht]
    \centering
    \caption{Specific value of the rank configuration explored by RankAdaptor in LLaMA-7B pruned by LLM-Pruner}
    \resizebox{0.9\linewidth}{!}
    {
        \begin{tabular}{cc|c}
            \toprule
             Pruning Rate &Tasks& Layers' Rank Values (1\textasciitilde 32/40) \\
            \midrule
              \multirow{7}{*}{20\%} & BoolQ & 8, 8, 8, 8, 4, 12, 12, 12, 12, 10, 8, 10, 6, 2, 8, 6, 8, 2, 8, 2, 8, 10, 12, 10, 6, 4, 4, 4, 2, 12, 8, 8\\
             \cmidrule{2-3}
             &PIQA&8, 8, 8, 8, 4, 12, 12, 12, 12, 10, 8, 10, 6, 2, 8, 6, 8, 2, 8, 2, 8, 10, 12, 10, 6, 4, 4, 4, 2, 12, 8, 8\\
             \cmidrule{2-3}
             &Hella&8, 8, 8, 8, 2, 2, 4, 10, 10, 6, 10, 10, 10, 6, 6, 2, 2, 10, 2, 4, 2, 10, 10, 10, 4, 10, 10, 6, 6, 2, 8, 8\\
             \cmidrule{2-3}
             &Wino&8, 8, 8, 8, 8, 10, 4, 10, 4, 6, 6, 2, 10, 8, 12, 12, 10, 12, 12, 10, 6, 6, 8, 8, 10, 6, 6, 12, 2, 8, 8, 8\\
             \cmidrule{2-3}
             &ARC-e&8, 8, 8, 8, 4, 12, 12, 12, 12, 10, 8, 10, 6, 2, 8, 6, 8, 2, 8, 2, 8, 10, 12, 10, 6, 4, 4, 4, 2, 12, 8, 8\\
             \cmidrule{2-3}
             &ARC-c&8, 8, 8, 8, 4, 12, 12, 12, 12, 10, 8, 10, 6, 2, 8, 6, 8, 2, 8, 2, 8, 10, 12, 10, 6, 4, 4, 4, 2, 12, 8, 8\\
             \cmidrule{2-3}
             &OBQA&8, 8, 8, 8, 4, 12, 12, 12, 12, 10, 8, 10, 6, 2, 8, 6, 8, 2, 8, 2, 8, 10, 12, 10, 6, 4, 4, 4, 2, 12, 8, 8\\
            \cmidrule{1-3}
             \multirow{7}{*}{25\%} & BoolQ & 8, 8, 8, 8, 12, 2, 8, 2, 8, 12, 4, 2, 10, 12, 10, 4, 2, 2, 12, 8, 10, 2, 12, 12, 8, 4, 4, 2, 2, 12, 8, 8\\
             \cmidrule{2-3}
             &PIQA&8, 8, 8, 8, 4, 2, 2, 10, 10, 2, 10, 10, 10, 2, 2, 2, 4, 10, 4, 6, 10, 2, 2, 6, 10, 2, 2, 10, 10, 2, 8, 8\\
             \cmidrule{2-3}
             &Hella&8, 8, 8, 8, 4, 10, 12, 12, 6, 10, 6, 6, 8, 2, 2, 12, 2, 12, 12, 6, 4, 10, 6, 2, 2, 8, 4, 2, 2, 8, 8, 8\\
             \cmidrule{2-3}
             &Wino&8, 8, 8, 8, 8, 4, 12, 8, 2, 2, 12, 2, 10, 12, 2, 12, 12, 10, 8, 12, 4, 6, 6, 4, 10, 4, 2, 10, 10, 12, 8, 8\\
             \cmidrule{2-3}
             &ARC-e&8, 8, 8, 8, 2, 12, 2, 6, 12, 6, 12, 10, 6, 4, 8, 8, 12, 2, 2, 6, 8, 4, 12, 12, 2, 4, 2, 6, 6, 2, 8, 8\\
             \cmidrule{2-3}
             &ARC-c&8, 8, 8, 8, 2, 12, 2, 6, 12, 6, 12, 10, 6, 4, 8, 8, 12, 2, 2, 6, 8, 4, 12, 12, 2, 4, 2, 6, 6, 2, 8, 8\\
             \cmidrule{2-3}
             &OBQA&8, 8, 8, 8, 4, 12, 12, 12, 12, 1, 8, 10, 6, 2, 8, 6, 8, 2, 8, 2, 8, 10, 12, 12, 10, 4, 4, 6, 2, 12, 8, 8\\
            \cmidrule{1-3}
              \multirow{7}{*}{30\%} & BoolQ & 8, 8, 8, 8, 12, 2, 8, 2, 8, 12, 4, 2, 10, 12, 10, 4, 2, 2, 12, 8, 10, 2, 12, 12, 8, 4, 4, 2, 2, 12, 8, 8\\
             \cmidrule{2-3}
             &PIQA&8, 8, 8, 8, 12, 6, 10, 4, 2, 4, 2, 4, 12, 8, 2, 2, 2, 12, 12, 12, 12, 2, 12, 4, 4, 2, 10, 2, 2, 8, 8, 8\\
             \cmidrule{2-3}
             &Hella&8, 8, 8, 8, 12, 6, 8, 4, 2, 12, 10, 4, 4, 2, 6, 4, 6, 10, 4, 2, 8, 6, 12, 10, 4, 6, 6, 6, 8, 2, 8, 8\\
             \cmidrule{2-3}
             &Wino&8, 8, 8, 8, 12, 6, 8, 4, 2, 12, 10, 4, 4, 2, 6, 4, 6, 10, 4, 2, 8, 6, 12, 10, 4, 6, 6, 6, 8, 2, 8, 8\\
             \cmidrule{2-3}
             &ARC-e&8, 8, 8, 8, 12, 6, 8, 4, 2, 12, 10, 4, 4, 2, 6, 4, 6, 10, 4, 2, 8, 6, 12, 10, 4, 6, 6, 6, 8, 2, 8, 8\\
             \cmidrule{2-3}
             &ARC-c&8, 8, 8, 8, 4, 10, 12, 12, 6, 10, 6, 6, 8, 2, 2, 12, 2, 12, 12, 6, 4, 10, 6, 2, 2, 8, 4, 2, 2, 8, 8, 8\\
             \cmidrule{2-3}
             &OBQA&8, 8, 8, 8, 12, 6, 8, 4, 2, 12, 10, 4, 4, 2, 6, 4, 6, 10, 4, 2, 8, 6, 12, 10, 4, 6, 6, 6, 8, 2, 8, 8\\
            \cmidrule{1-3}
             \multirow{7}{*}{50\%} & BoolQ & 8, 8, 8, 8, 12, 4, 6, 2, 2, 10, 4, 6, 12, 12, 2, 2, 12, 6, 4, 2, 6, 2, 4, 2, 6, 10, 10, 4, 2, 2, 8, 8\\
             \cmidrule{2-3}
             &PIQA&8, 8, 8, 8, 12, 4, 6, 2, 2, 10, 4, 6, 12, 12, 2, 2, 12, 6, 4, 2, 6, 2, 4, 2, 6, 10, 10, 4, 2, 2, 8, 8\\
             \cmidrule{2-3}
             &Hella&8, 8, 8, 8, 12, 4, 6, 2, 2, 10, 4, 6, 12, 12, 2, 2, 12, 6, 4, 2, 6, 2, 4, 2, 6, 10, 10, 4, 2, 2, 8, 8\\
             \cmidrule{2-3}
             &Wino&8, 8, 8, 8, 12, 4, 6, 2, 2, 10, 4, 6, 12, 12, 2, 2, 12, 6, 4, 2, 6, 2, 4, 2, 6, 10, 10, 4, 2, 2, 8, 8\\
             \cmidrule{2-3}
             &ARC-e&8, 8, 8, 8, 12, 4, 6, 2, 2, 10, 4, 6, 12, 12, 2, 2, 12, 6, 4, 2, 6, 2, 4, 2, 6, 10, 10, 4, 2, 2, 8, 8\\
             \cmidrule{2-3}
             &ARC-c&8, 8, 8, 8, 12, 4, 6, 2, 2, 10, 4, 6, 12, 12, 2, 2, 12, 6, 4, 2, 6, 2, 4, 2, 6, 10, 10, 4, 2, 2, 8, 8\\
             \cmidrule{2-3}
             &OBQA&8, 8, 8, 8, 12, 4, 6, 2, 2, 10, 4, 6, 12, 12, 2, 2, 12, 6, 4, 2, 6, 2, 4, 2, 6, 10, 10, 4, 2, 2, 8, 8\\
            \bottomrule
        \end{tabular}
    }\label{tab: con-llama7b}
\end{table*}
In Table \ref{tab: res-llama7b}, we present the performance achieved by RankAdaptor on each task. Table \ref{tab: con-llama7b} displays the rank configurations corresponding to the performance results of LLaMA-7B pruned by LLM-Pruner. We make these rank configurations publicly available to foster reproducibility and enable further research by other scholars.

\subsection{Generation Performance.}\label{Generative}
\begin{figure*}[h]
    \centering
    \includegraphics[width=\linewidth,bb=0 0 1100 350]{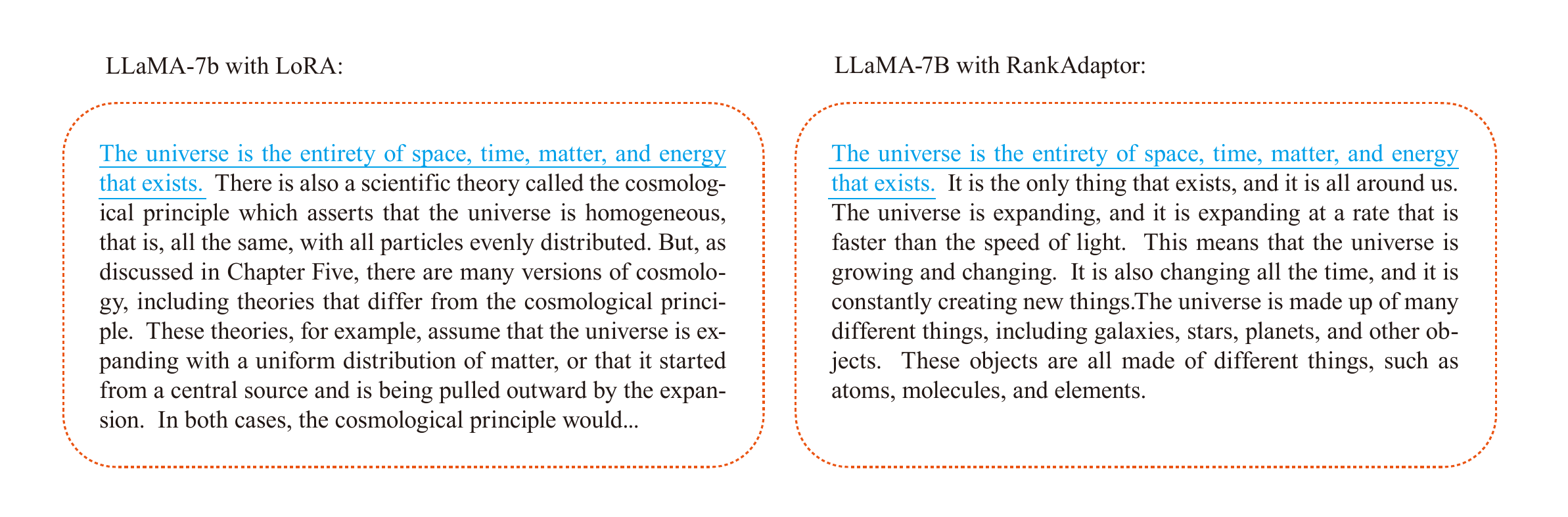}
    \caption{
        Article continuation task comparison in LLaMA-7B
    }
    \label{fig: gen1}
\end{figure*}

\begin{figure*}[t!]
    \centering
    \includegraphics[width=\linewidth,bb=0 0 1100 350]{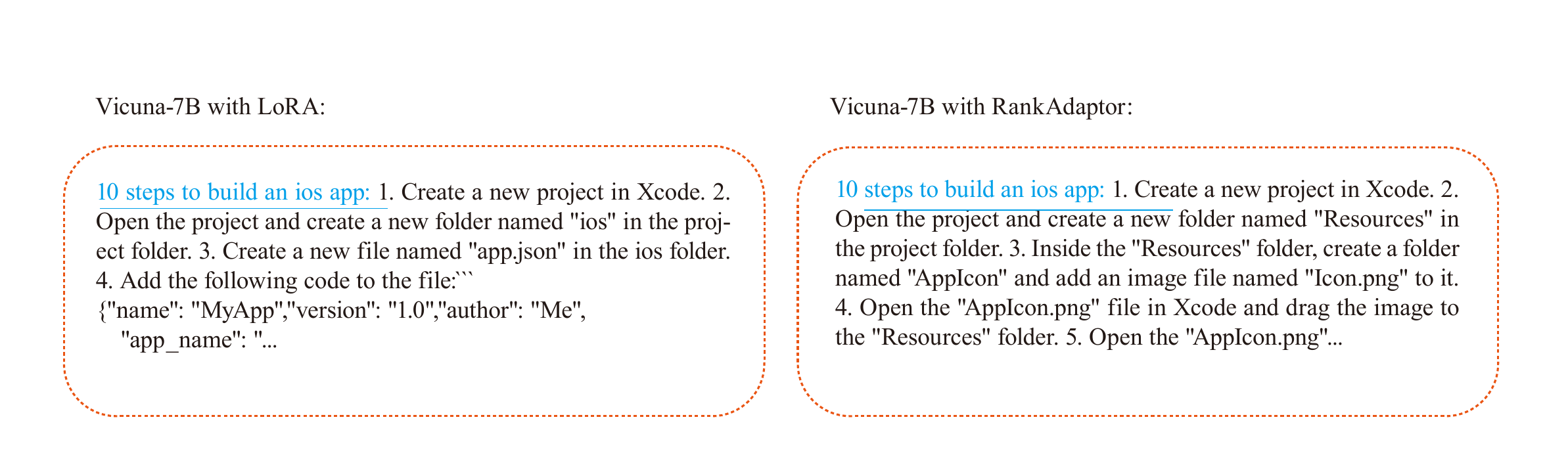}
    \caption{
        Step listing task comparison in Vicuna-7B
    }
    \label{fig: gen2}
\end{figure*}

The analysis of Generation Performance has been presented in Section \ref{sec: gen} of the main section. Therefore, here, we focus solely on presenting the generation comparison results between the pruned LLaMA-7B recovered by LoRA and RankAdaptor, as illustrated in Figures \ref{fig: gen1} and \ref{fig: gen2}.

\end{document}